\pdfoutput=1
\documentclass[11pt]{article}

\usepackage[final]{acl}

\usepackage{times}
\usepackage{latexsym}

\usepackage[T1]{fontenc}
\usepackage[utf8]{inputenc}

\usepackage{microtype}

\usepackage{inconsolata}

\usepackage{tikz}
\usepackage{tabularx}
\usepackage{booktabs}
\usepackage{multirow}
\usepackage{xspace}
\usepackage{amsmath}
\usepackage{cleveref}
\usepackage{amsfonts}
\usepackage{graphicx}
\usepackage{subcaption}
\usepackage{enumitem}

\newcommand{\ours}{\textsc{LLM2}\xspace}
\usepackage{colortbl}
\PassOptionsToPackage{table}{xcolor} % 统一加载 xcolor 包
\usepackage{xcolor}

\title{\ours: Let Large Language Models Harness System 2 Reasoning~\thanks{The work described in this paper is partially supported by a grant from the Research Grant Council of the Hong Kong Special Administrative Region, China (Project Code: 14200620).}}

\author{Cheng Yang$^{1,2}$\hspace{-1mm}~\thanks{Equal Contribution. This paper was completed during Cheng Yang’s time at Tsinghua University.}~~Chufan Shi$^{2\dag}$~~Siheng Li$^{1\dag}$~~Bo Shui$^{2}$~~Yujiu Yang$^{2}$~~Wai Lam$^{1}$\\ 
  $^{1}$The Chinese University of Hong Kong\quad $^{2}$Tsinghua University\\
  \href{mailto:yangc21@mails.tsinghua.edu.cn}{\texttt{yangc21@mails.tsinghua.edu.cn}} \\
  \textbf{Correspondence:} \href{mailto:sihengli24@gmail.com}{\texttt{sihengli24@gmail.com}}\quad\href{mailto:yang.yujiu@sz.tsinghua.edu.cn}{\texttt{yang.yujiu@sz.tsinghua.edu.cn}} \\
}

\begin{document}
\maketitle

\begin{abstract}
Large language models (LLMs) have exhibited impressive capabilities across a myriad of tasks, yet they occasionally yield undesirable outputs.
We posit that these limitations are rooted in the foundational autoregressive architecture of LLMs, which inherently lacks mechanisms for differentiating between desirable and undesirable results.  
Drawing inspiration from the dual-process theory of human cognition, we introduce LLM2, a novel framework that combines an LLM (System 1) with a process-based verifier (System 2). 
Within LLM2, the LLM is responsible for generating plausible candidates, while the verifier provides timely process-based feedback to distinguish desirable and undesirable outputs. The verifier is trained with a pairwise comparison loss on synthetic process-supervision data generated through our token quality exploration strategy. Empirical results on mathematical reasoning benchmarks substantiate the efficacy of LLM2, exemplified by an accuracy enhancement from 50.3 to 57.8 (+7.5) for Llama3-1B on GSM8K. Furthermore, when combined with self-consistency, \ours achieves additional improvements, boosting major@20 accuracy from 56.2 to 70.2 (+14.0)\footnote{Code is available at \href{https://github.com/yc1999/LLM2}{https://github.com/yc1999/LLM2}.}.
\end{abstract}
\section{Introduction}
\begin{figure*}[t]
    \centering
    % \vspace{-2.5em}
    \includegraphics[width=\linewidth]{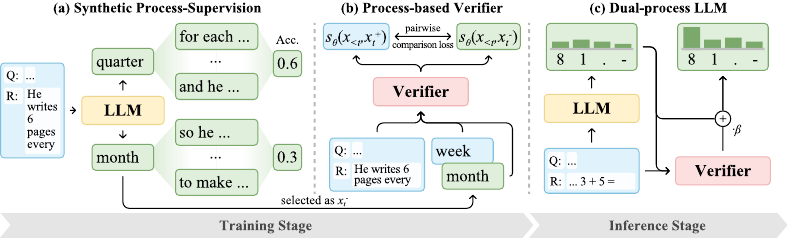}
    \caption{An illustration of the training and inference stages of LLM2. The training stage includes (a) synthetic process-supervision data collection and (b) the optimization of a process-based verifier. The inference stage involves (c) a dual-process LLM for generation.}
    \vspace{-1em}
    \label{fig:main}
\end{figure*}
Large language models \citep{brown2020language, chowdhery2023palm, gpt4} have exhibited remarkable abilities across various tasks that span general assistance \citep{chatgpt}, coding \citep{chen2021evaluating}, vision \citep{alayrac2022flamingo} and more. However, they still occasionally produce undesirable outputs in many scenarios, e.g., reasoning and planning \citep{mialon2023gaia, hu2023language}, factual consistency \citep{DBLP:conf/emnlp/MinKLLYKIZH23}, and human value alignment \citep{bai2022training}, etc. We hypothesize these deficiencies stem from the fundamental design of LLMs. Specifically, the next-token prediction objective optimizes LLMs to maximize the probability of human-generated strings empirically, with no explicit mechanism to distinguish between desirable and undesirable outputs. During the inference stage, LLMs autoregressively generate outputs token-by-token in a single pass, with no awareness of their errors. This procedure is reminiscent of System 1 in the dual-process theory, which postulates that thinking and reasoning are underpinned by two distinct cognitive systems \cite{stanovich200024, evans2003two, kahneman2011thinking}. System 1 operates automatically and subconsciously, guided by instinct and experience. In contrast, System 2, thought to be unique to humans, is more controlled and rational, enabling deliberate thinking for difficult tasks, especially when System 1 may make mistakes \citep{sloman1996empirical}.

In this paper, we introduce \ours, which aims to empower LLMs with System 2 reasoning. As shown in Figure \ref{fig:main}, \ours integrates an LLM (System 1) with a process-based verifier (System 2). During inference, the LLM generates multiple candidates at each time step, and the verifier provides timely feedback on each candidate. By efficiently exploring the generation space based on the verifier's feedback, \ours ultimately identifies more effective outputs. During the training stage, the process-based verifier is optimized with a pairwise comparison loss to distinguish between desirable and undesirable tokens. To obtain informative token pairs data for process-supervision, we propose a token quality exploration strategy that generates synthetic data based on the potential impact of tokens on the generated text.

We evaluate \ours on two representative mathematical reasoning datasets: GSM8K \cite{cobbe2021training} and MATH \cite{hendrycks2measuring}. With the integration of System 2 reasoning, \ours achieves substantial performance improvement across Llama3 models ranging from 1B to 8B parameters. For instance, compared to the vanilla Llama3-1B, \ours significantly improves accuracy from 50.3 to 57.8 (+7.5) on GSM8K, and from 24.2 to 28.8 (+4.6) on MATH. Combining \ours with self-consistency further boosts the model's performance, enhancing major@20 accuracy from 56.2 to 70.2 (+14.0) on GSM8K. Further analysis of the utilization of self-generated answers underscores the effectiveness and promising potential of synthetic process-supervision data.

\section{Method}

\subsection{Dual-process LLM}
\label{sec:dual_process_llm}
We aim to build a dual-process LLM (i.e., LLM2), where an LLM serves as System 1 for giving plausible proposals and a verifier functions as System 2 for deliberate thinking to refine and prevent mistakes introduced by System 1.
Specifically, we formalize this procedure as:
\begin{equation}
\scalebox{0.90}{
$\log \pi^* (x_t|x_{<t}) \propto \log \pi(x_t|x_{<t}) + \beta s(x_{<t}, x_t),$
}
\label{eq:dual_process}
\end{equation}
where $\pi$ and $\pi^*$ represent the policies of the LLM and dual-process LLM, respectively.
The verifier steers $\pi$ during decoding based on the process score $s(x_{<t}, x_t)$, with $\beta$ controlling the strength.
% For efficiency, we focus on the most probable tokens for verification at each time step, which are the top-$k$ tokens predicted by $\pi$, denoted as $\mathcal{V}_t(k)$. Accordingly,
For computational efficiency, we focus verification on the most probable tokens at each time step. 
Therefore, we filter out low probability tokens using an adaptive plausibility constraint \citep{li2022contrastive}:
\begin{equation}
\mathcal{V}_t = \{v \in \mathcal{V}: \mathbf{z}_t [v] \geq \log \alpha + \max_w \mathbf{z}_t [w] \},
\label{eq:constraint}
\end{equation}
where $\mathbf{z}_t$ represents the logits of $\pi$,
$\mathcal{V}$ is the vocabulary and $\mathcal{V}_t \subset \mathcal{V}$ denotes the token set filtered with the hyperparameter $\alpha \in [0,1]$ at time step $t$.

Therefore, the logits of $\pi^*$ at time step $t$, denoted as $\mathbf{z}^*_t$, are computed as:
\begin{equation}
    \mathbf{z}^*_t [v]=\begin{cases}\mathbf z_t[v] + \beta s(x_{<t}, v) &\text{if } v \in \mathcal{V}_t, \\
    -\infty & \text{otherwise}.
    \end{cases}
\end{equation}
The probability distribution $\pi^*(x_t|x_{<t})=\text{softmax}(\mathbf{z}^*_t)$.
This formulation allows $\pi^*$ to integrates seamlessly with various decoding strategies, depending on the use case.
\begin{figure*}[t]
    % \vspace{-2em}
    \centering
    \includegraphics[width=\linewidth]{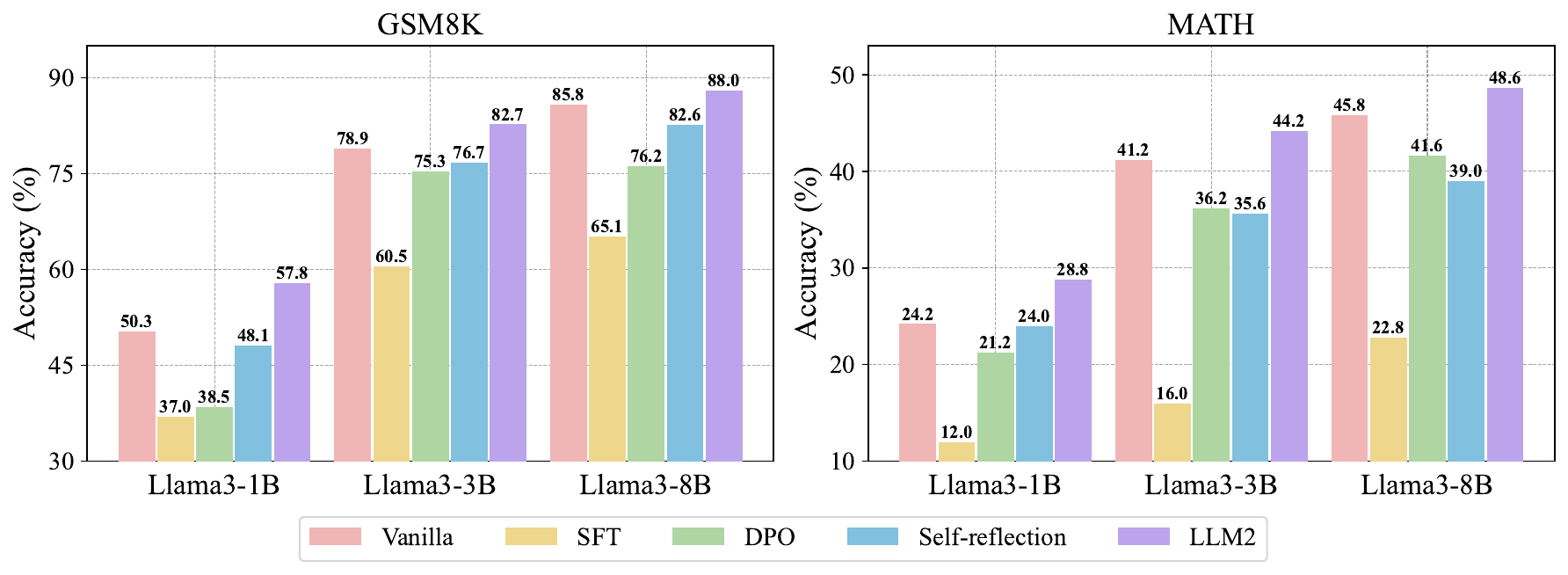}
    \vspace{-1em}
    \caption{Results of \ours and other baselines' performance on GSM8K and MATH with Llama3 series.}
    % \vspace{-1em}
    \label{fig:main_results}
\end{figure*}
\subsection{Process-based Verifier}
\label{sec:process_based_verifier}
We initialize the verifier from an LLM, replacing the unembedding head with a linear head to produce scalar scores.
Given a dataset $\mathcal{D}=\big\{x^{i}\big\}_{i=1}^N$, we synthesize process-supervision $\mathcal{D}_p(x) = \big\{x_{<t}, x_t^+, x_t^-\big\}_{t=1}^T$ for each instance $x$, where $x_t^+$ is more appropriate than $x_t^-$.
Accordingly, the training dataset for the verifier is $\mathcal{D}_s=\big\{x^{i}, \mathcal{D}_p(x^{i})\big\}_{i=1}^N$.
We train the verifier with a pairwise comparison loss \citep{ouyang2022training}:
\begin{align}
    \mathcal{L}&(s_\theta, \mathcal{D}_s) = -\mathbb{E}_{\big(x, \mathcal{D}_p(x)\big) \sim D_s} \nonumber\\
    & \sum_{t=1}^T \left[\log \sigma\big(s_{\theta}(x_{<t}, x_t^+) - s_{\theta}(x_{<t}, x_t^-)\big)\right].
\end{align}

\subsection{Synthetic Process-supervision}
\label{sec:synthetic_process_supervision}
We aim to create $\mathcal{D}_p(x) = \big\{x_{<t}, x_t^+, x_t^-\big\}_{t=1}^T$ for each instance $x$. 
In particular, we use the ground-truth token $x_t$ as $x_t^+$, which is desirable to be correct.
Regarding $x_t^-$, our goal is to select tokens that express the undesirable failure modes of LLMs, e.g., reasoning errors, hallucinations and misalignment with human values.
Then, through learning to distinguish between $x_t^+$ and $x_t^-$, the verifier can discern desirable and undesirable behaviors.

To create $x_t^-$, one can sample tokens from the distributions predicted by LLMs. However, LLMs may assign a high probability to alternative correct tokens, which leads to false $x_t^-$ and confuses the training of the verifier. To alleviate this issue, we introduce a token quality exploration strategy for sampling $x_t^-$. Specifically, the token quality exploration strategy evaluates the quality of individual tokens based on their potential impact on the generated text. This strategy involves three key steps:
\paragraph{Continuation Generation} For each candidate token $v \in \mathcal{V} \setminus \{x_t^{+}\}$ at time step $t$, we use the LLM to generate $N$ continuations $\{c_j\}_{j=1}^N$, each starting with $x_{<t}$ concatenated with $v$.
\paragraph{Quality Assessment} We evaluate the quality of each continuation based on the correctness of all decoded answers.
\begin{equation}
    q(v) = \frac{1}{N}\sum_{j=1}^N \text{quality}(c_j),
    \label{eq:quality_assessment}
\end{equation}
where $\text{quality}(c_j)$ is a function that returns the quality score for each continuation. In this work, we use accuracy as the quality measure.
\paragraph{Negative Sampling} We sample $x_t^-$ from tokens with low quality scores:
\begin{equation}
    x_t^- \sim \{v : q(v) < \tau, v \in \mathcal{V}_t \setminus \{x_t\}\},
    \label{eq:negative_sampling}
\end{equation}
where $\tau$ is a threshold hyperparameter.

The token quality exploration strategy enables the identification of tokens likely to lead to low-quality outputs, providing informative negative examples for training the verifier. In this work, we consider the top-$k$ most probable tokens according to the LLM's distribution as a candidate set, which reduces the computational cost while still capturing the most relevant candidates for $x_t^-$.

\section{Experiments}
\label{sec:experiments}

\subsection{Experimental Setup}

Our experiments are based on the Llama3 model series, specifically using 1B, 3B and 8B instruct versions~\citep{dubey2024llama}. We leverage these LLMs as System 1 and utilize them to initialize corresponding verifiers. We use the GSM8K training set as $\mathcal{D}$, and employ the LLMs to generate corresponding synthetic datasets $\mathcal{D}_{s}$ for training verifiers. For evaluation, we utilize two benchmarks: GSM8K \citep{cobbe2021training} and MATH \citep{hendrycks2measuring}. Further details regarding our experimental setup can be found in Appendix~\ref{appx:setting}.

\subsection{Results}
We present a comprehensive comparison of LLM2 against standard vanilla models and various pivotal baselines, including Self-reflection prompting \citep{madaan2024self}, Supervised Fine-tuning (SFT), and Direct Preference Optimization (DPO)~\citep{rafailov2024direct}. Further elaborations on these baselines are available in Appendix \ref{appx:baselines}.
As depicted in Figure \ref{fig:main_results}, implementing self-reflection prompting to engage the model in System 2 reasoning does not yield performance enhancements, suggesting a prevailing limitation in self-reflective capabilities for Llama3 models across different scales (1B, 3B, and 8B). Given that Llama3 has undergone extensive post-training with meticulously curated mathematical reasoning data~\cite{dubey2024llama}, applying GSM8K for either SFT or DPO training results in performance degradation across both GSM8K and MATH benchmarks. Conversely, \ours emerges as an effective approach to enhance Llama3's performance across different model size. Llama3-1B exhibits an increase from 50.3 to 57.8 (+7.5) on GSM8K, while Llama3-8B progresses from 85.8 to 88.0 (+2.2). Moreover, \ours demonstrates robust generalization capabilities, with improvements on MATH despite the process-based verifier's training on GSM8K. Specifically, Llama3-1B rises from 24.2 to 28.8 (+4.6) on MATH, and Llama3-8B advances from 45.8 to 48.6 (+2.6).

\section{Analysis}
\begin{table}[t]
\centering
\resizebox{\linewidth}{!}{
\begin{tabular}{c|c|cc}
\toprule
&  & \multicolumn{2}{c}{LLM2}         \\
\cmidrule(lr){3-4}
\multirow{-2}{*}{Task}  & \multirow{-2}{*}{Vanilla} & \quad \textit{w/} Ground Truth & \quad \textit{w/} SA\\
\midrule
GSM8K & 50.3  & 57.8 (+7.5) & \textbf{59.7} (+9.4) \\
MATH  & 24.2  & 28.8 (+4.6) & \textbf{30.2} (+6.0) \\
\bottomrule
\end{tabular}
}
\caption{
Results of using ground truth or self-generated answers~(SA) for \ours's synthetic process-supervision on GSM8K and MATH using Llama3-1B.
}
\label{tab:synthetic}
\end{table}
\subsection{Self-generated Answers for Synthetic Process-supervision}
We further refine our methodology by utilizing the model's self-generated correct answers as $\mathcal{D}$, replacing traditional golden solutions to formulate $\mathcal{D}_{s}$ for training verifiers. Instances that remain incorrect after multiple samplings are excluded. Our experiments with Llama3-1B, as illustrated in Table~\ref{tab:synthetic} indicate that crafting $\mathcal{D}$ from self-generated data enhances the efficacy of \ours. On GSM8K, performance heightens from 57.8 to 59.7, marking an improvement of 9.4 over the vanilla model. On MATH, results improve from 28.8 to 30.2, signifying a 6.0 increase over the baseline.

\subsection{Self-consistency}
We investigate the potential of integrating \ours with self-consistency~\cite{wang2022self}, with detailed setup provided in Appendix~\ref{appx:sc}. 
As demonstrated in Figure~\ref{fig:sc}, experiments conducted on Llama3-1B unveil that LLM2, when amalgamated with self-consistency, notably enhances performance. LLM2 trained with self-generated data (i.e., LLM2-SA) elevates Major@20 accuracy on GSM8K from 56.2 to 72.2, and on MATH, the Major@20 accuracy improves from 32.8 to 37.0.
\begin{figure}[t]
    \vspace{-1em}
    \centering
    \includegraphics[width=\linewidth]{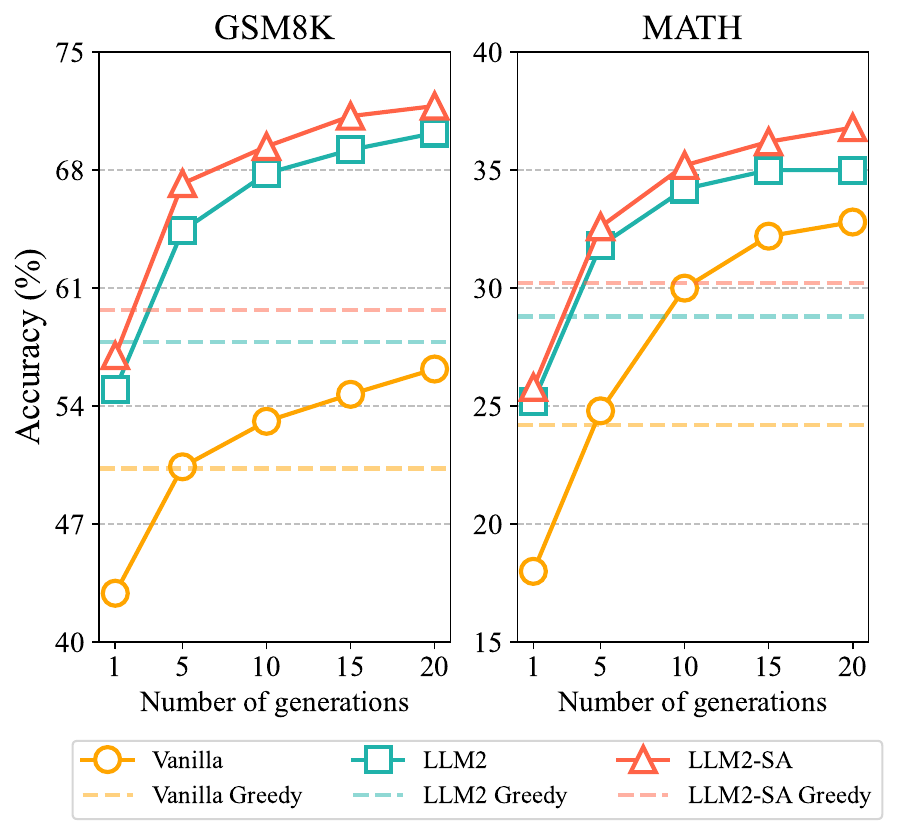}
    % \vspace{-2em}
    \caption{Results on combining \ours with self-consistency on GSM8K and MATH using Llama3-1B.}
    \label{fig:sc}
\end{figure}

% \begin{table}[htbp]
\begin{table}[t]
% \vspace{-2em}
\centering
\resizebox{\linewidth}{!}{
\begin{tabular}{l|ccc}
\toprule
\multirow{2}{*}{Method} & \multicolumn{3}{c}{Latency} \\
\cmidrule(lr){2-4}
 & \textsc{1B} & \textsc{3B} & \textsc{8B} \\
\midrule 
\textsc{Vanilla} & 2.8 ($\times$ 1.00) & 4.8 ($\times$ 1.00) & 5.3 ($\times$ 1.00) \\ 
\quad \textit{w/} \ours & 3.5 ($\times$ 1.25) & 5.9 ($\times$ 1.23) & 6.4 ($\times$ 1.21) \\ 
\bottomrule
\end{tabular}
}
\caption{Averaged per-instance decoding latency of \ours in seconds (s/example) on GSM8K.
}
\label{tab:latency}
\end{table}
\subsection{Latency}
We assess the impact of \ours's decoding latency and compare it with vanilla models on the Llama3 model series. Specifically, as shown in Table~\ref{tab:latency}, we report the averaged per-instance inference latency on GSM8K. Since the process-based verifier in \ours only performs inference when the LLM provides multiple candidate tokens after the adaptive plausibility constraint, \ours introduces an additional 1.21x to 1.25x latency. This latency tends to decrease as the modes's parameters increase.
\begin{table}[t]
\centering
% \resizebox{\linewidth}{!}{
\small
\begin{tabular}{c|cccc|c}
\toprule
& \multicolumn{4}{c|}{Math-Shepherd (Best-of-N)} & \\
\cmidrule(lr){2-5}
\multirow{-2}{*}{Task} & 5 & 10 & 15 & 20 & \multirow{-2}{*}{LLM2} \\
\midrule
GSM8K & 51.6 & 54.4 & 56.0 & 57.6 & \textbf{59.7} \\
MATH  & 26.4 & 27.2 & 27.0 & 27.0 & \textbf{30.2} \\
\bottomrule
\end{tabular}
% }
\caption{
Performance comparison between Math-Shepherd (Best-of-$N$)~\cite{wang2024math} and LLM2 on GSM8K and MATH using Llama3-1B. 
}
\label{tab:prm}
\end{table}
\subsection{Comparison with PRM Method}
We compare LLM2 with Math-Shepherd~\cite{wang2024math}, a representative Process Reward Model~(PRM) baseline for Llama3-1B, with the results presented in Table~\ref{tab:prm}. For a fair comparison, we use the GSM8K subset\footnote{\url{https://huggingface.co/datasets/peiyi9979/Math-Shepherd}} to train a Llama3-1B PRM model as the baseline. The results show that Math-Shepherd's performance converges at Best-of-$N$ ($N$=20), achieving 57.6 and 27.0 on GSM8K and MATH, respectively, while LLM2 achieves 59.7 and 30.2, demonstrating LLM2's advantages. Additionally, using PRM's Best-of-$N$ for inference potentially introduces an $N$-fold latency, whereas LLM2 only incurs approximately 1.2x latency. This demonstrates the advantage of LLM2's token-level supervision signals~\cite{lin2024critical}, which enable more efficient and precise optimization during the generation process.
\begin{table}[t]
\centering
\resizebox{\linewidth}{!}{
\begin{tabular}{c|cccc|c}
\toprule
Task & Vanilla & SFT & DPO & Self-reflection & LLM2 \\
\midrule
GSM8K & 69.2 & 56.0 & 60.3 & 68.7 & \textbf{73.5} (+4.3)\\
MATH  & 46.4 & 22.8 & 38.6 & 43.8 & \textbf{49.0} (+2.6)\\
\bottomrule
\end{tabular}
}
\caption{
    Results of \ours and other baselines' performance on GSM8K and MATH with Qwen2.5-1.5B.
}
\label{tab:qwen}
\end{table}
\subsection{Employ Qwen2.5}
We further investigate the generalizability of LLM2 across diverse LLM families, conducting experiments on the Qwen2.5-1.5B model~\cite{qwen2.5}. As illustrated in Table~\ref{tab:qwen}, LLM2 emerges as a robust approach to enhance the performance of Qwen2.5-1.5B on both the GSM8K and MATH benchmarks. Specifically, compared to the vanilla model, LLM2 achieves notable improvements in mathematical reasoning, with performance gains of 4.3 and 2.6 on GSM8K and MATH, respectively. In contrast, other methods fail to surpass the vanilla baseline, highlighting the unique efficacy of LLM2. This aligns with our observations on the Llama3 model series, where LLM2 consistently enhanced performance across different model sizes and tasks, reinforcing its potential as a universal enhancement framework for different LLM families.

\section{Related Work}
\paragraph{Verifier for LLMs.} Training verifiers to explicitly distinguish between desirable and undesirable outputs has been a promising method to improve the capabilities of LLMs. Existing verifier modeling can be broadly classified into two categories: (1) Outcome-based modeling \citep{DBLP:conf/emnlp/ShenYLSJ0021, cobbe2021training}, which train verifiers to learn how to distinguish between correct and wrong outputs and selects more optimal ones from a number of candidates at inference time. (2) Process-based modeling~\citep{DBLP:journals/corr/abs-2211-14275, DBLP:journals/corr/abs-2305-20050,zhu2023solving}, which supervises each reasoning step of the generation process. 
To alleviate the reliance on human-annotated process-supervision data, \citet{wang2024math} propose to automatically construct process-supervision data, where the correctness of a mathematical reasoning step is defined as its potential to reach the final answer correctly.

In LLM2, we propose a process-based verifier to emulate System 2 reasoning. It is trained on synthetic process-supervision data generated by our token quality exploration strategy. During inference, this verifier can intervene at any time step, providing immediate feedback without waiting for the completion of specific steps or the entire output.

\paragraph{System 2 for LLMs.} Recent works explore the incorporation of System 2 into LLMs, primarily during the inference stage~\citep{weston2023system,deng2023rephrase,saha2024branch}. These approaches often leverage System 2 mechanisms, such as reflection and planning~\citep{madaan2024self}, to generate explicit and verbalized reasoning content, which then guides subsequent token generation. Alternatively, some research focuses on transferring System 2 capabilities to System 1 during the training phase through methods such as distillation~\citep{yu2024distilling}, thereby obviating the need for generating intermediate reasoning tokens during the inference stage.

LLM2 integrates System 2 during the inference stage. Specifically, LLM2 leverages a process-based verifier as System 2 to provide real-time feedback at each token generation step without generating auxiliary content.

\section{Conclusion}
In this work, we introduce \ours, a framework that augments LLMs with a System 2-like reasoning process. By coupling an LLM with a process-based verifier, LLM2 proficiently differentiates between optimal and suboptimal outputs. The framework is empowered by synthetic process-supervision data generated via a novel token quality exploration strategy, which is instrumental in training the verifier. Our empirical results and analyses confirm the efficacy of LLM2 in enhancing LLM performance.

\section*{Limitations}
While \ours demonstrates significant improvements in mathematical reasoning tasks, our exploration does not extend to other reasoning domains, such as commonsense reasoning and code generation, due to computational resource constraints. We are optimistic about the potential of \ours to generalize well to these additional tasks. However, applying \ours to open-ended tasks, like creative writing, presents challenges due to the lack of definitive supervisory signals for synthetic process-supervision. Addressing these challenges offers a promising direction for future research.

\section*{Acknowledgments}
This work was partly supported by the National Key Research and Development Program of China (No. 2024YFB2808903), the research grant No. CT20240905126002 of the Doubao Large Model Fund and the Shenzhen Science and Technology Program JSGG20220831110203007).

% Bibliography entries for the entire Anthology, followed by custom entries
%\bibliography{anthology,custom}
% Custom bibliography entries only
\bibliography{custom}

\clearpage

\appendix

% \clearpage
\section{Experimental Setup}
\label{appx:setting}
\paragraph{Dataset.} We leverage the training set of GSM8K \citep{cobbe2021training} as $\mathcal{D}$ and use the test set of GSM8K as one of our evaluation set. Although we do not use the MATH \citep{hendrycks2measuring} train set to train the verifier, we utilize the MATH test set as an additional evaluation set to validate the effectiveness of the verifier in improving general mathematical reasoning. Due to computational resource constraints, we randomly sampled 500 examples from the original MATH test set for our evaluation.
\paragraph{Hyperparameter Setting.} We generally set $\beta$ to 0.25 in Equation \ref{eq:dual_process}, $\alpha$ to 0.1 in Equation \ref{eq:constraint} and $\tau$ to 0.5 in Equation \ref{eq:negative_sampling}. We set $N$ to 20 in Equation \ref{eq:quality_assessment}. For top-$k$ in Section \ref{sec:synthetic_process_supervision}, $k$ is set to 5.

\paragraph{Model Details.} We list the Llama3 and Qwen2.5 models used in our experiments along with their corresponding HuggingFace model names in Table \ref{tab:model_card}.
\begin{table}[ht]
\resizebox{\linewidth}{!}{\begin{tabular}{lr}
\toprule
Model     & HuggingFace Model Name                \\ \midrule
Llama3-1B    & \texttt{meta-llama/Llama-3.2-1B-Instruct}      \\
Llama3-3B &  \texttt{meta-llama/Llama-3.2-3B-Instruct}\\
Llama3-8B & \texttt{meta-llama/Llama-3.1-8B-Instruct} \\
Qwen2.5-1.5B & \texttt{Qwen/Qwen2.5-1.5B-Instruct} \\
\bottomrule
\end{tabular}}
\caption{Llama 3 and Qwen2.5 models and their corresponding HuggingFace model names.}
\label{tab:model_card}
\end{table}
\paragraph{Details of Training Verifiers.} We train our verifiers using 8 NVIDIA A100 80GB GPUs. The training process is conducted over 3 epochs with a batch size of 128. We employ a learning rate of 2e-5 and utilize a cosine learning rate scheduler.
% \section{Details of Self-generated Answers for Synthetic Process-supervision}
% \label{appx:SA}
\section{Baselines}
\label{appx:baselines}
We implement four representative baselines:
\paragraph{Vanilla} utilizes the original Llama model directly for inference.
\paragraph{Supervised Fine-tuning (SFT)} fine-tunes LLMs to maximize the log-likelihood of the training data, which in our case is the GSM8K training set. The training process is conducted over 3 epochs with a batch size of 128. We employ a learning rate of 2e-5 and utilize a cosine learning rate scheduler.
\paragraph{Direct Preference Optimization (DPO)} \citep{rafailov2024direct} optimizes language models directly from desirable and undesirable outputs, eliminating the need for an explicit reward model. For desirable data, we use the GSM8K training set; for undesirable data, a randomly sampled incorrect output from the model serves as the undesirable example. The training process is conducted over 1 epoch with a batch size of 128. We set ${\beta}=0.01$ and employ a learning rate of 5e-7 and utilize a cosine learning rate scheduler.
\paragraph{Self-reflection Prompting}~\citep{madaan2024self} involves first generating an output, followed by prompting the model to assess whether its output is correct and whether to revise the output. This approach can be seen as introducing System 2 reasoning through prompting. The specific prompt is shown in Table \ref{tab:prompt}. 
\newcolumntype{Y}[1]{%
  >{\small\everypar{\hangindent=1em}\arraybackslash}p{#1}%
}

\begin{table}[h]
\small
\begin{tabular}{@{}Y{0.95\linewidth}@{}}
\toprule
% \texttt{\textbf{Genre:} Science Fiction}\\
% \midrule
\texttt{Please review your answer. If you think it is correct, just repeat your answer. If you think it is incorrect, please generate the correct one.}\\

\bottomrule
\caption{Prompt for Self-reflection prompting.}
\vspace{-2em}
\label{tab:prompt}
\end{tabular}
\end{table}
\section{Self-consistency Setup}
\label{appx:sc}
For vanilla self-consistency, we use temperature sampling with temperature $\tau=1.0$ for instruct models to reach the best baseline performance~\cite{shi2024thorough}. For combining \ours with self-consistency, we simply set $\beta$ to 0.25 in Equation \ref{eq:dual_process}, $\alpha$ to 0.1 in Equation \ref{eq:constraint} and do temperature sampling with temperature $\tau=1.0$.

\section{Comparison with Token-Level Decoding Methods}
To further demonstrate the effectiveness of our process-based verifier, we compare LLM2 with token-level decoding methods. Specifically, we implement contrastive decoding (CD)~\citep{li2022contrastive} and DoLa~\citep{chuang2023dola}, and evaluate their performance on the GSM8K and MATH datasets. The results are shown in Tables~\ref{tab:decoding_gsm8k} and \ref{tab:decoding_math}.

For CD, we follow the hyperparameter settings from \citet{li2022contrastive,o2023contrastive,shi2024unchosen}, using Llama3-1B as the amateur model. For DoLa, we follow the hyperparameter settings from \citet{chuang2023dola,shi2024thorough}. The results reported for both CD and DoLa represent their best performance across their hyperparameter ranges. As shown, CD does not yield significant improvements, primarily because CD requires an ideal amateur model~\citep{o2023contrastive,shi2024thorough} which may not always exist. As for DoLa, while it proves effective for factual knowledge tasks, it can have adverse effects on reasoning tasks~\citep{chuang2023dola,shi2024thorough}.

\begin{table}[h]
\centering
\begin{tabular}{lcccc}
\toprule
Model & Vanilla & CD & DoLa & LLM2 \\
\midrule
Llama3-1B & 50.3 & - & 47.2 & \textbf{57.8} \\
Llama3-3B & 78.9 & 79.8 & 76.1 & \textbf{82.7} \\
Llama3-8B & 85.8 & 86.4 & 83.0 & \textbf{88.0} \\
\bottomrule
\end{tabular}
\caption{Results of token-level decoding methods on GSM8K with Llama3 series.}
\label{tab:decoding_gsm8k}
\end{table}

\begin{table}[h]
\centering
\begin{tabular}{lcccc}
\toprule
Model & Vanilla & CD & DoLa & LLM2 \\
\midrule
Llama3-1B & 24.2 & - & 23.6 & \textbf{28.8} \\
Llama3-3B & 41.2 & 42.0 & 39.6 & \textbf{44.2} \\
Llama3-8B & 45.8 & 46.4 & 43.2 & \textbf{48.6} \\
\bottomrule
\end{tabular}
\caption{Results of token-level decoding methods on MATH with Llama3 series.}
\label{tab:decoding_math}
\end{table}

\begin{table}[t]
\centering
\resizebox{\linewidth}{!}{
\begin{tabular}{c|ccc}
\toprule
Model & Llama3-1B & Llama3-3B & Llama3-8B \\
\midrule
Accuracy & 75.5 & 77.0 & 79.4 \\
\bottomrule
\end{tabular}
}
\caption{
    Accuracy of \ours verifier~(1B, 3B and 8B) on GSM8K for the corresponding Llama3 model series.
}
\label{tab:verifier}
\end{table}
\section{Accuracy of Process-based Verifier}
We further analyze the accuracy of LLM2's process-based verifier in distinguishing between ground-truth and non-ground-truth tokens. Specifically, using the GSM8K test set, we pair each question $q$ with its answer $a$. Then we leverage the vanilla models to perform next-token prediction tasks on $(q,a_{<t})$ and collect the non-ground-truth token with the highest probability as $\tilde{a_{t}}$. Subsequently, we input $(q, a_{<t}, a_{t})$ and $(q, a_{<t}, \tilde{a_{t}})$ into the corresponding verifier. A correct prediction is determined by whether the verifier assigns a higher score to $(q, a_{<t}, a_{t})$. The results, presented in Table~\ref{tab:verifier}, demonstrate the verifier's effective token-level accuracy.

\section{Case Study}
\begin{table*}[htp]
\setlength\tabcolsep{3pt}
\scriptsize
\centering
    \begin{tabular} {p{1.0cm}|p{14cm}}
        \toprule
         \textbf{Question} & Lloyd has an egg farm. His chickens produce 252 eggs per day and he sells them for \$2 per dozen. How much does Lloyd make on eggs per week?
         \\
         \midrule
         \multirow{12}{*}{\textbf{Vanilla}} & \#\# Step 1: Calculate the number of eggs produced per week
         
Since there are 7 days in a week, Lloyd's chickens produce 252 eggs per day * 7 days = 1754 eggs per week.

\\
& \#\# Step 2: Convert the number of eggs per week to dozens

Since there are 12 eggs in a dozen, Lloyd produces 1754 eggs / 12 = 146.8333 dozens of eggs per week.

\\
& \#\# Step 3: Calculate the total amount of money Lloyd makes per week

Lloyd sells each dozen of eggs for \$2, so he makes 146.8333 dozens * \$2 = \$292.6666 per week.

\\
& \#\# Step 4: Round the total amount of money Lloyd makes per week to the nearest dollar

Rounding \$292.6666 to the nearest dollar gives \$293.

\\
& The final answer is: \text{\$\textbackslash boxed\{293\}\$}\\
         \midrule
         \multirow{12}{*}{\textbf{LLM2}} & \#\# Step 1: Calculate the number of eggs produced per week

To find out how many eggs Lloyd produces per week, we need to multiply the number of eggs produced per day by 7. eggs\_per\_day = 252, eggs\_per\_week = eggs\_per\_day * 7 = 252 * 7 = 1764.

\\
& \#\# Step 2: Calculate the number of dozens of eggs produced per week

Since there are 12 eggs in a dozen, we need to divide the total number of eggs produced per week by 12 to find out how many dozens of eggs Lloyd produces per week. dozens\_per\_week = eggs\_per\_week / 12 = 1764 / 12 = 147.

\\
& \#\# Step 3: Calculate the total amount of money Lloyd makes per week

To find out how much Lloyd makes per week, we need to multiply the number of dozens of eggs produced per week by the price per dozen. price\_per\_dozen = \$2, total\_money\_per\_week = dozens\_per\_week * price\_per\_dozen = 147 * 2 = 294.

\\
& The final answer is: \$\textbackslash boxed\{294\}\$\\
         \bottomrule
    \end{tabular}
    \caption{A case study from GSM8K using Llama3-1B, where LLM2 corrects the vanilla model's arithmetic error.}
    \label{tab:case1}
\end{table*}

\begin{table*}[htp]
\setlength\tabcolsep{3pt}
\scriptsize
\centering
    \begin{tabular} {p{1.0cm}|p{14cm}}
        \toprule
         \textbf{Question} &  Terry eats 2 yogurts a day.  They are currently on sale at 4 yogurts for \$5.00.  How much does he spend on yogurt over 30 days?
         \\
         \midrule
         \multirow{5}{*}{\textbf{Vanilla}} & \#\# Step 1: Calculate the cost per yogurt
         
The cost per yogurt is \$5.00 / 4 yogurts = \$1.25 per yogurt.

\\
& \#\# Step 2: Calculate the total cost for 30 days

The total cost for 30 days is \$1.25 per yogurt * 30 days = \$37.50.

\\
& The final answer is: \$\textbackslash boxed\{37.5\}\$ \\
         \midrule
         \multirow{7}{*}{\textbf{LLM2}} & \#\# Step 1: Calculate the cost of one yogurt

To find out how much Terry spends on yogurt per day, we need to calculate the cost of one yogurt. The current sale price of 4 yogurts for \$5.00 is \$5.00 / 4 = \$1.25 per yogurt.

\\
& \#\# Step 2: Calculate the cost of yogurt over 30 days

Now that we know the cost of one yogurt, we can calculate the total cost of yogurt over 30 days. Terry eats 2 yogurts a day, so over 30 days, he eats 2 * 30 = 60 yogurts. The total cost of yogurt over 30 days is 60 * \$1.25 = \$75.00.

\\
& The final answer is: \$\textbackslash boxed\{75\}\$ \\
         \bottomrule
    \end{tabular}
    \caption{A case study from GSM8K using Llama3-1B, where LLM2 corrects the vanilla model's logical error.}
    \label{tab:case2}
\end{table*}
We present two representative cases from GSM8K using Llama3-1B to demonstrate how LLM2 improves mathematical reasoning in Table~\ref{tab:case1} and \ref{tab:case2}.

In Case 1, LLM2 demonstrates its ability to prevent computational errors. While the vanilla model made an arithmetic error in calculating weekly egg production (252 × 7 = 1754), LLM2 correctly computed 1764 eggs per week, leading to the accurate final answer of 294.

In Case 2, LLM2 shows how it prevents logical errors. The vanilla model overlooked Terry's daily consumption of 2 yogurts, while LLM2 correctly accounted for both the unit price (1.25) and total consumption (60 yogurts over 30 days), yielding the correct answer of 75.

These cases demonstrate how LLM2's verification mechanism helps maintain both computational and logical accuracy throughout the reasoning process.

\end{document}